\documentclass{article}

\usepackage{arxiv}

\usepackage[utf8]{inputenc} 
\usepackage[T1]{fontenc}    
\usepackage{url}            
\usepackage{booktabs}       
\usepackage{amsfonts}       
\usepackage{nicefrac}       
\usepackage{microtype}      
\usepackage{lipsum}		
\usepackage{graphicx}
\usepackage{natbib}
\usepackage{doi}
\usepackage{amsmath}

\usepackage{multirow}
\usepackage{makecell}

\usepackage{hyperref}       

\title{MIPHEI-ViT: Multiplex Immunofluorescence Prediction from H\&E Images using ViT Foundation Models}

\date{} 					

\author{
\href{https://orcid.org/0000-0002-7750-8852}{\includegraphics[scale=0.06]{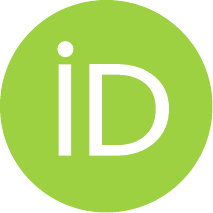}\hspace{1mm}Guillaume Balezo} \\
Digital R\&D, Sanofi, \\
Center for Computational Biology (CBIO) Mines Paris - PSL, \\
Center for Statistics and Images (STIM) Mines Paris - PSL \\
Paris, 75008, France \\
\texttt{guillaume.balezo@minesparis.psl.eu} \\
\And
Roger Trullo\thanks{Present address: InstaDeep, 75009 Paris, France.} \\
Digital R\&D, Sanofi \\
Paris, 75008, France \\
\texttt{r.trullo@instadeep.com} \\
\And
\href{https://orcid.org/0000-0001-7527-2500}{\includegraphics[scale=0.06]{orcid.pdf}\hspace{1mm}Albert Pla Planas} \\
Digital R\&D, Sanofi, \\
Barcelona, 08016, Spain \\
\texttt{albert.plaplanas@sanofi.com} \\
\And
\href{https://orcid.org/0000-0002-1349-8042}{\includegraphics[scale=0.06]{orcid.pdf}\hspace{1mm}Etienne Decencière} \\
Center for Statistics and Images (STIM) Mines Paris - PSL \\
Fontainebleau, 77300, France \\
\texttt{etienne.decenciere@minesparis.psl.eu} \\
\And
\href{https://orcid.org/0000-0001-7419-7879}{\includegraphics[scale=0.06]{orcid.pdf}\hspace{1mm}Thomas Walter}\thanks{Corresponding author: thomas.walter@minesparis.psl.eu} \\
Center for Computational Biology (CBIO) Mines Paris - PSL, \\
Institut Curie, INSERM, U1331\\
Paris, 75005, France \\
\texttt{thomas.walter@minesparis.psl.eu} \\
}

\renewcommand{\headeright}{}
\renewcommand{\undertitle}{Accepted Manuscript}
\renewcommand{\shorttitle}{MIPHEI-ViT: Multiplex Immunofluorescence Prediction from H\&E Images}

\hypersetup{
pdftitle={MIPHEI-ViT: Multiplex Immunofluorescence Prediction from H\&E Images using ViT Foundation Models},
pdfsubject={q-bio.NC, q-bio.QM},
pdfauthor={Guillaume B., Roger T., Albert P., Etienne D., Thomas W.},
pdfkeywords={Computer Vision, Histopathology, Image Translation, Foundation model, In silico labelling},
}

\begin{document}
\maketitle

\begin{center}
Published in \textit{Computers in Biology and Medicine} (2026), Open Access.\\
DOI: \href{https://doi.org/10.1016/j.compbiomed.2026.111564}{10.1016/j.compbiomed.2026.111564}\\
\vspace{0.3em}
© 2026. Licensed under CC BY-NC-ND 4.0:
\href{https://creativecommons.org/licenses/by-nc-nd/4.0/}{creativecommons.org/licenses/by-nc-nd/4.0}\\
\vspace{0.3em}
\textit{Preprint: 15 May 2025 \textbar{} Accepted: 19 February 2026}
\end{center}

\vspace{1em}

\begin{abstract}
Histopathological analysis is a cornerstone of cancer diagnosis, with Hematoxylin and Eosin (H\&E) staining routinely acquired for every patient to visualize cell morphology and tissue architecture. On the other hand, multiplex immunofluorescence (mIF) enables more precise cell type identification via proteomic markers, but has yet to achieve widespread clinical adoption due to cost and logistical constraints. 
To bridge this gap, we introduce MIPHEI (Multiplex Immunofluorescence Prediction from H\&E Images), a U-Net-inspired architecture that leverages a ViT pathology foundation model as encoder to predict mIF signals from H\&E images using rich pretrained representations. MIPHEI targets a comprehensive panel of markers spanning nuclear content, immune lineages (T cells, B cells, myeloid), epithelium, stroma, vasculature, and proliferation. We train our model using the publicly available OrionCRC dataset of restained H\&E and mIF images from colorectal cancer tissue, and validate it on five independent datasets: HEMIT, PathoCell, IMMUcan, Lizard and PanNuke. On OrionCRC test set, MIPHEI achieves accurate cell-type classification from H\&E alone, with F1 scores of 0.93 for Pan-CK, 0.83 for $\alpha$-SMA, 0.68 for CD3e, 0.36 for CD20, and 0.28 for CD68, substantially outperforming both a state-of-the-art baseline and a random classifier for most markers. Our results indicate that, for some molecular markers, our model captures the complex relationships between nuclear morphologies in their tissue context, as visible in H\&E images and molecular markers defining specific cell types. MIPHEI offers a promising step toward enabling cell-type-aware analysis of large-scale H\&E datasets, in view of uncovering relationships between spatial cellular organization and patient outcomes.
\end{abstract}


\section{Introduction}
\label{sec:introduction}
The analysis of Hematoxylin and Eosin (H\&E)-stained tissue slides is a cornerstone in the diagnosis of many pathologies, including cancer, providing insights into cell types, cellular phenotypes, tissue architecture and their alterations due to disease. 

Multiplex immunofluorescence (mIF) \citep{b1, b2, b3} imaging is a powerful technique that improves the analysis of tissue sections by providing detailed information beyond what H\&E staining can reveal. mIF achieves this by simultaneously visualizing and quantifying multiple protein markers within a single tissue section, utilizing fluorescently labeled antibodies that bind to specific proteins, allowing for the identification of cell types based on marker expression. This capability makes mIF useful across various domains, including cancer biology, immunology, and infectious disease, where understanding spatial cell organization is critical. By combining molecular and spatial information at single-cell resolution, mIF supports detailed characterization of tissue microenvironments and cellular interactions.

The evolution of immunolabeling techniques from traditional immunohistochemistry (IHC) \citep{b4} has led to more advanced mIF techniques like PhenoCycler \citep{b5}, which enables the detection of up to 100 markers through multiple imaging cycles, each capturing 4 channels. This iterative process can however degrade tissue integrity. Among recent advancements, the ORION scanner introduced by \citet{b6} allows the simultaneous detection of up to 20 markers in a single cycle, preserving tissue quality while still providing rich multiplexed information. ORION also allows capturing high-quality, restained H\&E images from the same tissue section.

While mIF imaging offers several advantages, it also presents important challenges. Preparing and processing mIF slides is time-consuming and labor-intensive, requiring expensive reagents and specialized equipment, which are not always available in all clinical settings, limiting its accessibility. Compared to other techniques for spatial biology, such as Imaging Mass Cytometry (IMC) \citep{b7} and VisiumHD spatial transcriptomics \citep{visiumhd}, mIF detects fewer molecular markers but offers much higher spatial resolution. Like these technologies, mIF is constrained by high costs, which prevents it from being adopted in clinical practice.

In contrast, H\&E slides are routinely generated in clinical practice. Since different cell types are characterized by distinct protein expression patterns, we hypothesize that certain markers can be predicted from cell morphology captured in H\&E. With the availability of high-quality datasets obtained from technologies like the ORION scanner, which captures aligned H\&E and mIF images with a high number of markers, we can now train AI models to infer mIF data from H\&E images. This would allow us to identify those proteins that are predictable from morphological cues and perform these predictions on large retrospective cohorts, including clinical trial data, for which an acquisition of mIF data would not be feasible. Relating the predicted mIF data to outcome or treatment response can then contribute to biomarker discovery and hypothesis generation in oncology. 

\begin{figure}[h]
\centering
\includegraphics[width=\textwidth]{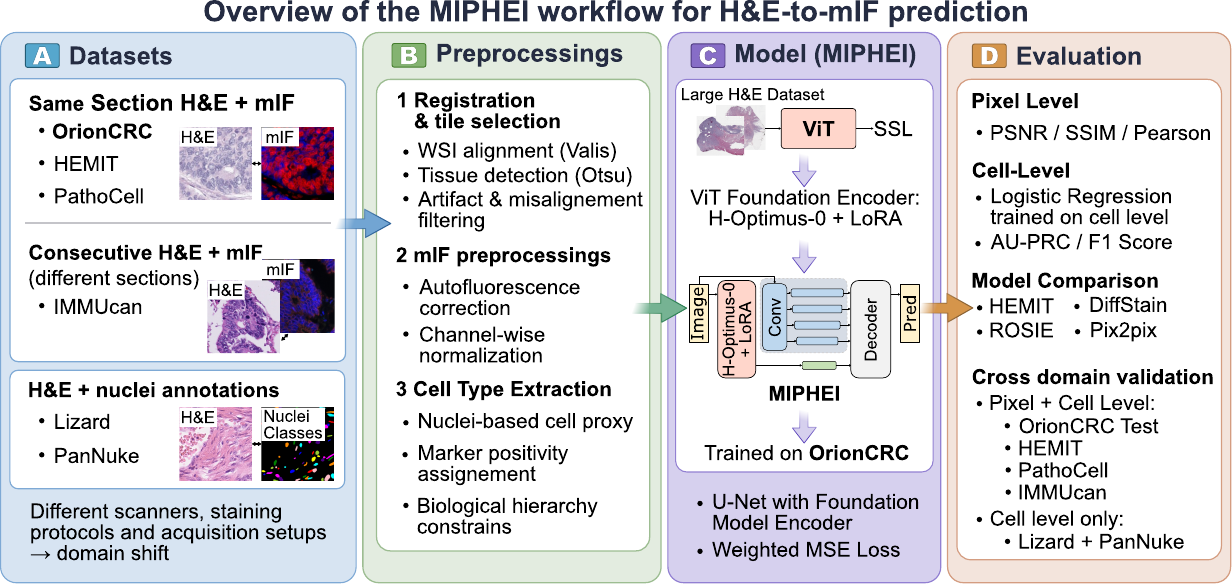}
\caption{\textbf{Overall workflow of the study.} \textbf{(A)} Datasets used for training and evaluation. \textbf{(B)} Data preprocessing, including alignment, tiling, mIF preprocessing, and cell-type extraction. \textbf{(C)} Our proposed MIPHEI architecture for H\&E-to-mIF prediction. \textbf{(D)} Pixel- and cell-level evaluation on in-domain and out-of-domain datasets.}
\label{figure_workflow}
\end{figure}

In this study, we aim to predict the expression of key markers from H\&E, covering nuclear content (Hoechst), vasculature (CD31), immune populations (CD45, CD68, CD4, FOXP3, CD8a, CD45RO, CD20, PD-L1, CD3e, CD163), epithelial and stromal structures (E-cadherin, Pan-CK, $\alpha$-SMA), and proliferation (Ki67). To achieve this, we introduce \textbf{MIPHEI}: \textbf{M}ultiplex \textbf{I}mmunofluorescence \textbf{P}rediction from \textbf{H}\&\textbf{E}-stained Whole Slide \textbf{I}mages, a U-Net-inspired architecture integrating state-of-the-art foundation models as encoders. Unlike traditional cell type classification models that rely on manual labels or pseudo-labels, our method is trained directly on aligned mIF data, avoiding biases from predefined annotations. It is also highly modular and does therefore not rely on specific nucleus segmentation and cell classification methods. An overview of the complete MIPHEI workflow is shown in Figure~\ref{figure_workflow}.

The key contributions of this study are as follows:

\begin{itemize}
    \item \textbf{ViT Foundation Models in U-Net Architecture}: one major methodological novelty is the integration of pathology Vision Transformer (ViT) \citep{b25} foundation models as the encoder in a U-Net architecture for histopathological image translation. This design benefits from rich representations learned from millions of diverse tiles across various staining protocols and sample preparations.
    \item \textbf{Prediction of Numerous Markers from H\&E}: We identify the proteomic markers and cell types predictable from H\&E images from a wide range of 15 markers.
    \item \textbf{Rigorous Validation and Reproducibility}: 
    We demonstrate generalization of the method through validation with cell-level metrics for accurate cell type identification on both internal and five external datasets, moving beyond pixel-level evaluation. We further provide a unified benchmark enabling direct comparison of multiple existing H\&E-to-mIF methods.
    \item \textbf{New State-of-the-art}: Our model outperforms previous models and a random baseline on 15 markers, most of which are indiscernible to pathologists.
    \item \textbf{Open Benchmark and Code}: We release the full pipeline, pretrained models, and datasets to support reproducibility and future research in H\&E to mIF prediction, available on \href{https://github.com/Sanofi-Public/MIPHEI-ViT}{GitHub}\footnote{\textbf{GitHub:}~\url{https://github.com/Sanofi-Public/MIPHEI-ViT}} and Zenodo \citep{miphei_dataset}, with an interactive demo on \href{https://huggingface.co/spaces/Estabousi/MIPHEI-vit-demo}{Hugging Face}\footnote{\textbf{Demo:}~\url{https://huggingface.co/spaces/Estabousi/MIPHEI-vit-demo}} (see Supplementary Video 1 for a demonstration of the interactive interface).

\end{itemize}

\section{Related Works}

\subsection{Virtual Staining}

Image-to-image translation involves transforming images from one domain to another, with Pix2Pix \citep{b8} being one of the most well-known methods in this domain for paired images. Utilizing conditional adversarial networks, these techniques enable high-quality image synthesis, supporting tasks such as style transfer, super-resolution, and domain adaptation, and have driven significant advancements across various applications, as demonstrated by \citet{b9}. More recently, diffusion-based generative models have gained significant attention for high-quality image generation, including models such as Stable Diffusion \citep{stablediffusion}.

In recent years, image-to-image translation techniques have become increasingly popular for life science applications, such as predicting super-resolved microscopy images \citep{b10} or bright-field-like images from holographic images \citep{b14}. In histopathology, such virtual staining techniques have been used to predict immunohistochemistry (IHC) data \citep{b11, b21}, to do virtual multiplexing \citep{b12, b13, Ounissi2025}, to predict mIF images from IHC images \citep{b15}, or both H\&E \citep{b16, b17, b18, b19} and mIF \citep{b20, zhou2024virtual, loo2025autofluorescence} from label-free imaging, such as autofluorescence or phase contrast microscopy.

In the scope of this study, the most relevant work involves translating H\&E images to mIF, which directly addresses the challenge of performing cell type calling from standard histological stains. Several approaches address this problem under different supervision regimes. Histoplexer \citep{histoplexer} predicts proteomic signals from H\&E using consecutive IMC tissue sections. Other methods instead rely on same-section H\&E–mIF pairs, which avoids imperfect registration and small variations in cellular content that can arise with consecutive sections, and therefore facilitates reliable cell-level supervision. A notable example is SHIFT \citep{b22} based on conditional generative adversarial network (cGAN) to predict markers like DAPI, pan-cytokeratin (Pan-CK), and $\alpha$-smooth muscle actin. HEMIT \citep{b24, b23} further introduces a public dataset together with a hybrid Convolutional Neural Network (CNN)–Transformer generator combining convolutional layers with Swin Transformers \citep{b31} for H\&E-to-mIHC translation, targeting DAPI, CD3, and Pan-CK markers. More recently, diffusion-based approaches have been explored, notably DiffusionFT \citep{diffusionft}, which builds on the Stable Diffusion \citep{stablediffusion} pipeline to enable efficient per-marker mIF prediction. Finally, beyond pixel-level image translation, ROSIE \citep{rosie} uses a ConvNeXt \citep{convnextv2} model to predict per-channel mean mIF signals within a small central region of H\&E images, yielding downsampled virtual staining prediction when applied in a sliding-window manner.

\subsection{ViT on Dense Prediction Tasks}

ViTs \citep{b25} are now widely used in image classification and often outperform CNNs when pretrained with advanced contrastive self-supervised learning (SSL) methods such as MoCov3 \citep{b30}, iBOT \citep{ibot}, and DINOv2 \citep{b28}. These SSL approaches have been adapted to histopathology, enabling recent foundation models to achieve strong and robust performance across diverse downstream tasks. CTransPath \citep{b29} uses MoCov3 \citep{b30} with a Swin Transfomer architecture \citep{b31}, and was trained on 15.6 million tiles from opensource datasets. UNIv2 \citep{b33} leverages DINOv2 to train a ViT-H/14 model on 200 million tiles from 3.5k proprietary H\&E and IHC slides. H-optimus-0 \citep{hoptimus} is a ViT-G/14 variant also trained with DINOv2 on hundreds of millions of tiles from 500,000+ H\&E whole slide images (WSIs). These models are among the state-of-the-art in computational pathology, with DINOv2-based models like UNIv2 and H-optimus-0 outperforming ImageNet-pretrained encoders and CTransPath on unseen datasets.

While plain ViTs are effective for image-level tasks, they often struggle with dense prediction tasks like image translation due to limited ability to capture fine local details, unlike CNNs which excel at leveraging local continuity and multiscale features. To overcome this, hybrid CNN-ViT architectures have been proposed. CellViT \citep{b40}, for example, uses the UNETR architecture \citep{b41}, which integrates a ViT encoder into a U-Net-like design, employing convolutional transpose blocks to create hierarchical features, offering a simple adaptation. ViTMatte \citep{vitmatte} combines a plain ViT with a convolutional module for pyramidal feature extraction and detail refinement, using ViT token features only as the bottleneck. More advanced models like ViT-Adapter \citep{vitadapter} and ViT-CoMer \citep{b42} go further by enabling richer multi-scale interactions between ViT and CNN components, but at higher computational cost. While these hybrid methods have mainly been applied to segmentation, we hypothesized they could also benefit dense image translation tasks like mIF prediction, especially when using foundation models trained on large, diverse datasets.

\section{Dataset}

\begin{table}[t]
\centering
\caption{Overview of the datasets used in this study. MPP denotes the resolution in microns per pixel. HEMIT images were downsampled by a factor of 2 to match the 20x resolution used in our experiments.}
\label{tab:datasets}
\footnotesize
\setlength{\tabcolsep}{4pt}
\begin{tabular}{llcccc}
\toprule
Dataset & Split & \#Tiles & Size (px) & Scanner & MPP \\
\midrule
\multicolumn{6}{l}{\textbf{Same-section H\&E--mIF datasets}} \\
\midrule
\multirow{3}{*}{OrionCRC}
 & Train & 295096 & 333$\times$333 & Aperio GT450 & 0.5 \\
 & Val   & 12402  & 333$\times$333 & Aperio GT450 & 0.5 \\
 & Test  & 10952  & 333$\times$333 & Aperio GT450 & 0.5 \\
HEMIT      & Test & 5292 & 512$\times$512 & Mantra & $\approx$0.5 \\
PathoCell  & Test & 4807 & 256$\times$256 & Pannoramic P250 & 0.377 \\
\midrule
\multicolumn{6}{l}{\textbf{Consecutive-section H\&E--mIF datasets}} \\
\midrule
IMMUcanCRC & Test & 16933 & 1024$\times$1024 & NanoZoomer 2.0-HT & 0.5 \\
\midrule
\multicolumn{6}{l}{\textbf{H\&E nuclei datasets}} \\
\midrule
PanNuke & Test & 7901 & 256$\times$256 & Multi & $\approx$0.25 \\
Lizard  & Test & 4777 & 256$\times$256 & Multi & $\approx$0.5 \\
\bottomrule
\end{tabular}
\end{table}

\begin{figure}[!t]
\centering
\includegraphics[width=\textwidth]{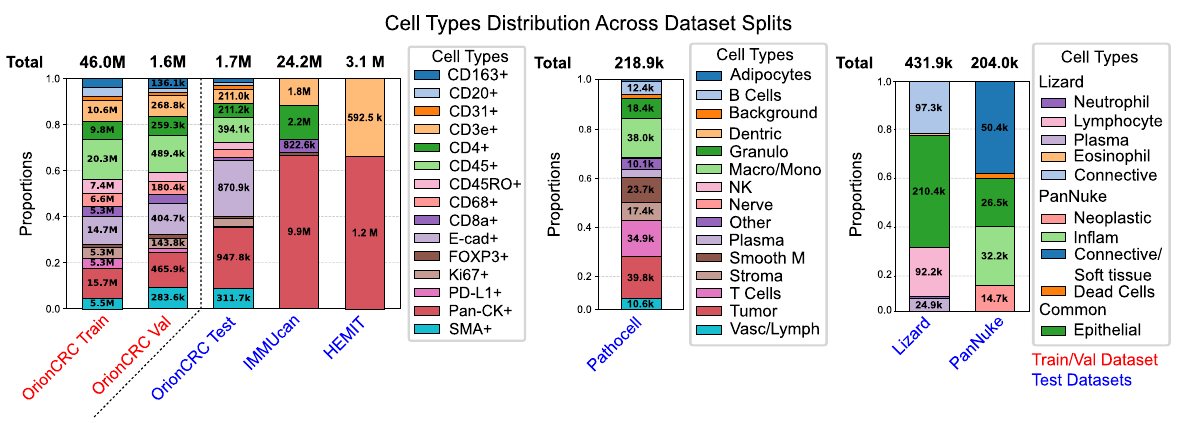}
\caption{\textbf{Cell-type distribution across datasets.} Proportions of cell types across datasets and splits. Datasets highlighted in red are used for training and validation. Cell-type labels vary between datasets, and rare cell types may appear at very low proportions.}
\label{figure_dataset}
\end{figure}

We present the datasets used in this study, with Table~\ref{tab:datasets} summarizing their composition and acquisition details, and Figure~\ref{figure_dataset} illustrating cell type distributions across datasets. The preprocessed data for OrionCRC and HEMIT datasets is available on \href{https://doi.org/10.5281/zenodo.15340874}{Zenodo}.

\subsection{OrionCRC Dataset (Train/Val/Test)}
The ORION colorectal cancer (OrionCRC) dataset \citep{orioncrc, b6} was acquired with a novel system capturing 18-channel immunofluorescence (IF) images alongside H\&E staining on the same tissue sections. This dataset includes 41 WSIs with both H\&E and mIF data. The 15 markers (with additional DNA stain) of interest for this study are: Hoechst, CD31, CD45, CD68, CD4, FOXP3, CD8a, CD45RO, CD20, PD-L1, CD3e, CD163, E-cadherin, Pan-CK, $\alpha$-SMA and Ki67. The PD-1 channel was not used due to poor signal quality.

\subsection{HEMIT Dataset (Test)}

The HEMIT dataset \citep{b24} consists of H\&E and multiplex-immunohistochemistry (mIHC) images the same tissue sections, with DAPI, CD3, and Pan-CK markers.

\subsection{PathoCell Dataset (Test)}

The PathoCell dataset was introduced in \cite{pathocell_original} and comprises 106 tissue microarrays (TMAs) cores obtained from around 70 colorectal cancer (CRC) specimens originating from 35 patients. Tissues were imaged using a 56-marker PhenoCycler \citep{b5} mIF panel together with same section H\&E. The dataset was later adopted in \cite{pathocell} to evaluate foundation models for cell phenotyping. In this setting, cell instances and cell phenotypes were derived from the mIF measurements (cf. Section~\ref{subsubsec:single-cell-clustering}), resulting in 15 categories. For this dataset, we applied only minimal preprocessing (cf Section~\ref{subsubsec:single-cell-clustering}), providing a clearly independent dataset for evaluation.

\subsection{IMMUcan CRC Dataset (Test)}

IMMUcan (Integrated iMMUnoprofiling of large adaptive CANcer patient cohorts) \citep{b43} is a European initiative launched in 2019 to advance Tumor Micro Environment (TME) profiling. We use only the CRC cases, comprising 35 registered H\&E and mIF WSIs from two cohorts, which include both advanced and less severe stages. mIF images were acquired on a PerkinElmer Vectra Polaris and the mIF panel includes DAPI, CD3, CD8, CD4, FOXP3, and Pan-CK. Unlike HEMIT, PathoCell and OrionCRC, the H\&E and mIF slides in IMMUcan come from consecutive sections, requiring a dedicated evaluation protocol detailed later in Section~\ref{subsection:cell_metrics}. Although currently private, the dataset is expected to become publicly available.

\subsection{Expert-annotated H\&E nuclei datasets (Test)}

To further evaluate MIPHEI at the cell level, we use two expert-annotated H\&E nuclei datasets: Lizard \citep{lizard}  and PanNuke \citep{pannuke}. These datasets provide ground-truth cell-type annotations directly from H\&E images and are used exclusively for out-of-domain evaluation. Lizard is a multi-center CRC cohort containing 291 fields of view. PanNuke spans 18 organs and provides expert annotations of nuclei into five cell categories. It is the only dataset in our evaluation that includes non-CRC tissue. For both datasets, cell-type labels are derived from expert annotations on H\&E images and therefore correspond to coarser categories than mIF-derived cell phenotypes.

\subsection{Data Preprocessing}

\begin{figure}[!t]
\centerline{\includegraphics[width=\textwidth]{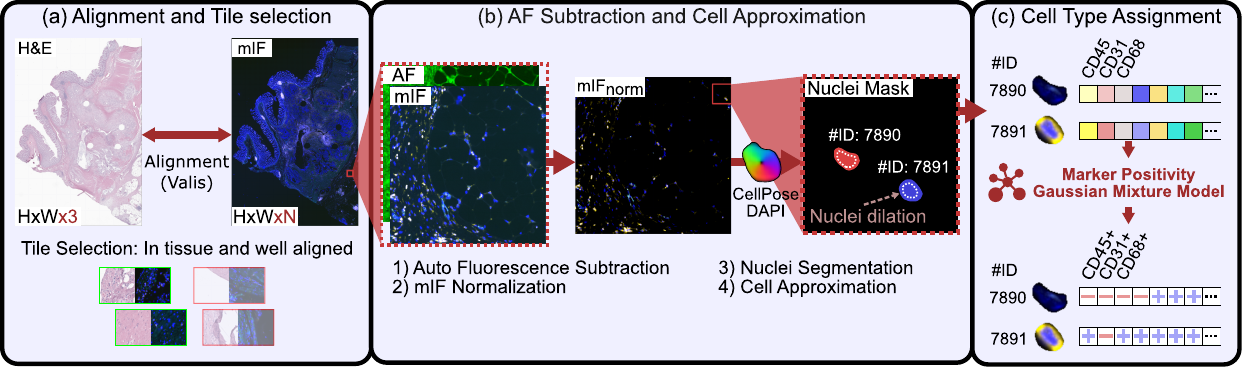}}
\caption{\textbf{Preprocessing pipeline:} \textbf{(a)} H\&E and mIF images are aligned using Valis \citep{valis}. Tissue regions are then selected via Otsu thresholding on H\&E, and a trained CNN filters misaligned tiles caused by the restaining and acquisition process. \textbf{(b)} Autofluorescence is subtracted from mIF images, followed by DAPI-based nuclei segmentation  and nuclei dilation to approximate cell boundaries. \textbf{(c)} Pseudo-labels are generated by computing per-cell mean marker expression and applying Gaussian Mixture Model (GMM) clustering to define marker positivity, determining labels (e.g., CD3e\textsuperscript{+}).}
\label{figure_pipeline}
\end{figure}

We designed a preprocessing pipeline tailored to our H\&E-to-mIF prediction task, addressing WSI registration, artifact removal, autofluorescence subtraction, normalization, and cell pseudo-label assignement, as illustrated in Figure \ref{figure_pipeline}.

\subsubsection{H\&E to mIF registration on consecutive cuts}

Accurate registration between H\&E and mIF images is crucial for training and evaluating our models. For OrionCRC, HEMIT and PathoCell datasets, data was already registered \citep{orioncrc, b24}. For the IMMUcan dataset, we used Valis \citep{valis} for registration of consecutive H\&E and mIF slides.

\subsubsection{Tile Selection}

HEMIT and PanNuke are provided as pre-extracted tiles. For OrionCRC, PathoCell, IMMUcan, and Lizard, tiles were extracted from tissue regions identified using Otsu thresholding, with tile sizes defined per dataset (Table~\ref{tab:datasets}).

To reduce artifacts from restaining and acquisition in the OrionCRC dataset, we filtered aligned H\&E and mIF tiles using several quality control steps. We used thresholding on an empty mIF channel—without associated antibody but capturing shared noise—to identify artifacts affecting all mIF channels. Poor H\&E quality tiles were identified by clustering H-optimus-0 embeddings and discarding clusters associated to obvious artifacts. We also manually annotated misaligned tiles and trained a CNN to detect them automatically from H\&E and DAPI images. In total, about 40k tiles (10\%) were excluded.

For the IMMUcan dataset, where H\&E and mIF are from consecutive sections, we selected well-aligned tiles to ensure reliable correlation analysis based on three criteria: Pearson correlation of aligned nuclei density maps from 32×32 patches (threshold=0.25 per tile), tissue overlap ($IoU> 0.5$), and  tissue percentage ($>40\%$). We extracted $1024\times 1024$ pixel tiles at 20x (0.26 mm$^2$), retaining 17k tiles covering 44.4 cm$^2$.

\subsubsection{Autofluorescence Subtraction \& Data Normalization}

Autofluorescence (AF), captured as an independent channel \( I_{AF} \), refers to light naturally emitted by the tissue across channels, introducing noise in other markers. This signal correlates with H\&E morphology and can artificially inflate reconstruction metrics if not accounted for. We subtract the AF from each marker channel \( I_{IF}^c \), using a linear subtraction with parameters \( \lambda^c \) and \( b^c \), which were manually adjusted using a Napari \citep{napari} tool we developed:

\begin{align}
I_{cor}^c = \max(0, I_{IF}^c - \lambda^c \cdot I_{AF} + b^c)
\end{align}

AF subtraction is applied exclusively to the OrionCRC dataset, as it is the only one containing an independent AF channel.

Next, each channel is normalized using the 99.9th percentile \( q_{0.999}^c \), computed per marker from the distribution of foreground pixel intensities (i.e., values > 0) across the training set, and then log-transformed:

\begin{align}
I_{norm}^c = 255 \cdot \log\left(\frac{\min(I_{cor}^c, q_{0.999}^c)}{q_{0.999}^c} + 1\right)
\end{align}

This logarithmic transformation compresses high-intensity values and reduces the impact of extreme outliers, while normalization ensures a consistent dynamic range across markers. This normalization is applied to OrionCRC and PathoCell, while HEMIT mIF is provided already normalized. The autofluorescence Napari tool, selected parameters, code, model, and data access instructions are available on our \href{https://github.com/Sanofi-Public/MIPHEI-ViT}{GitHub}, within the Sanofi Public organization, under specific license conditions including a limitation to non-commercial use only.

\subsubsection{Single-Cell Pseudo-Label Extraction}
\label{subsubsec:single-cell-clustering}

To establish a ground truth for evaluating our model’s ability to identify marker-positive cells from H\&E images, we used a standard cell type calling approach on mIF images \citep{b6}. We first segmented nuclei from the DAPI channel using Cellpose \citep{cellpose} fine-tuned on our data. As a proxy for cell regions, we dilated the nuclear regions by 2 $\mu m$. Single-cell analysis was performed by extracting per-channel mean fluorescence intensities, followed by unsupervised gating using a Gaussian Mixture Model (GMM) to distinguish positive from negative cells \citep{celesta}, with posterior probabilities estimated from the GMM. Although effective, this approach is sensitive to artifacts, approximate boundaries, and signal spillover. To improve robustness, we implemented a hierarchical gating strategy making sure that the biological marker hierarchy was preserved. For instance, given that CD3 positive cells are also CD45 positive, we kept CD3 positivity only for CD45 positive cells. By applying these rules, we obtained a high confidence annotation of our cells.

For the IMMUcan dataset, we used single-cell data provided by the consortium, which was generated using in-house clustering and nucleus segmentation. Since the same nuclei are not necessarily present in consecutive H\&E and mIF sections, we applied Hoverfast \citep{b44} for nucleus segmentation from H\&E images, allowing us to perform the cell-level correlation analysis explained in Section~\ref{subsec:metric_overview}.

For PathoCell, Lizard, and PanNuke, cell labels were taken directly from the original studies. All three datasets provide nuclei instance masks and cell-type annotations: labels are manually annotated for Lizard and PanNuke, while PathoCell labels are derived from mIF measurements.

\begin{figure}[!t]
\centering
\includegraphics[width=0.8\textwidth]{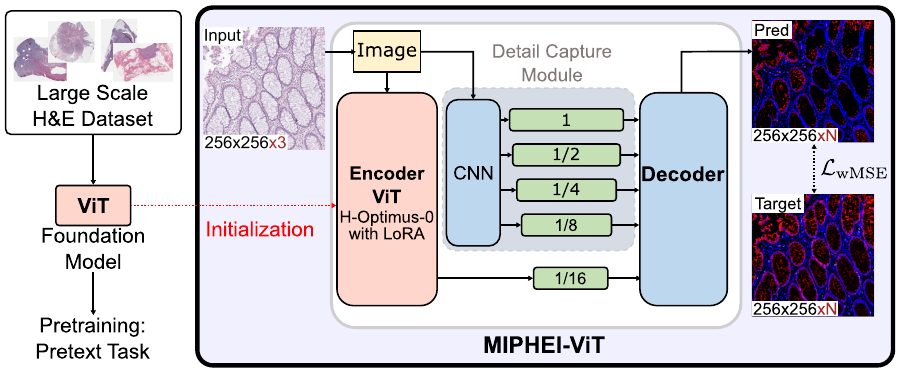}
\caption{\textbf{MIPHEI architecture:} A U-Net-inspired model based on VitMatte for H\&E-to-mIF prediction. The main contribution is the integration of a pretrained histology ViT foundation encoder (H-optimus-0) into an image translation model to leverage large-scale pretraining on diverse pathology images. A lightweight Detail Capture module complements the ViT encoder by injecting in the decoder missing spatial information for dense prediction in a U-Net manner. H-Optimus-0 uses a patch size of 14 and patch tokens are resized using bicubic interpolation to match the 1/16 spatial resolution required by the decoder. A Tanh activation is applied, with a custom weighted MSE loss to address marker imbalance. mIF images are shown as composite RGB images (Hoechst in blue, Pan-CK in red).}
\label{figure_architecture}
\end{figure}

\section{Methodology}

This Section outlines our overall approach for predicting mIF images from H\&E images. We used the OrionCRC dataset for training, and HEMIT and IMMUcan datasets for testing.

\subsection{Model Architecture}

MIPHEI is a U-Net-inspired architecture designed for image translation from H\&E to mIF signals (Figure \ref{figure_architecture}). The framework supports various encoder types, including CNNs like ConvNeXt v2 \citep{convnextv2}, Swin Transformer \citep{b31}, and plain ViTs \citep{b25}, enabling integration of recent foundation models in histology, such as CTransPath, Univ2, and H-optimus-0 \citep{b29, b33, hoptimus}. When using CNN-based encoders, MIPHEI follows a standard U-Net formulation, as these encoders naturally provide pyramidal features. For ViT-based encoders, which maintain fixed feature dimensions across layers, MIPHEI adopts a ViTMatte-inspired \citep{vitmatte} hybrid design: a lightweight convolutional Detail Capture Module extracts multiscale pyramidal features, while ViT features are used as the bottleneck. This module enables the extraction of fine-grained spatial information for dense prediction in a U-Net manner. Unlike original ViTMatte, which concatenates a trimap to the input image,  MIPHEI operates directly on the H\&E RGB image and omits convolutional necks and window attention mechanisms. Based on the analysis presented in Section~\ref{subsubsec:fmbenchmark}, we selected H-optimus-0 as the ViT encoder for MIPHEI.

Building on these encoded features, the decoder reconstructs outputs using bilinear interpolation for upsampling, combined with skip connections, a 3×3 convolutional layer, batch normalization, and ReLU activation. Each mIF marker is predicted by a dedicated output head applied to the decoder output, followed by a Tanh activation.

We also define a MIPHEI variant following UNETR \citep{b41} desing, which employs convolutional transpose operations to upsample ViT token features extracted at different depths. While this approach achieves performance comparable to the ViTMatte-inspired design, the choice of the ViTMatte-based formulation is discussed in Section~\ref{subsubsec:architecturecomparison}.

\subsection{Loss Function}

We train MIPHEI using a weighted Mean Square Error (MSE) loss to ensure accurate translation from H\&E to mIF signals while accounting for varying signal intensities and prevalence across markers. Each marker’s loss is scaled by the inverse of its standard deviation to balance contributions \citep{weightedmse}.

Let \( L_{\text{MSE}, j} \) denote the MSE loss for the \( j^{\text{th}} \) marker, \( \sigma_j \) denote the standard deviation, \( M \) be the total number of markers, and \( \lambda \) the weight of the reconstruction loss. The weighted MSE loss is then defined as:

\begin{align}
\mathcal{L}_{\text{wMSE}} = \frac{\lambda}{M} \sum_j \frac{1}{\sigma_j} L_{\text{MSE}, j}
\end{align}

\subsection{Foundation Model Training Strategies}
\label{subsec:finetuning-strats}

While the CNN-based encoders within the U-Net architecture are fully fine-tuned in all experiments, we explore two efficient training strategies for large ViT-based foundation model encoders: (1) decoder-only training with a frozen foundation encoder, leveraging robust pretrained features while reducing trainable parameters; (2) Low-Rank Adaptation (LoRA) \citep{lora} with \textit{rank} = 8 and $\alpha = 1$, which adapts a pretrained ViT encoder by adding trainable low-rank matrices to the Query and Value projections of the attention layers. Both strategies are common fine-tuning practices for large foundation models when training data is limited to low- or medium-sized datasets.

\section{Experiments}
\label{sec:experiments}

Here we present the experimental workflow that we set up to assess the performance and robustness of our image translation model from H\&E to mIF.

\begin{figure}[!t]
\centerline{\includegraphics[width=\textwidth]{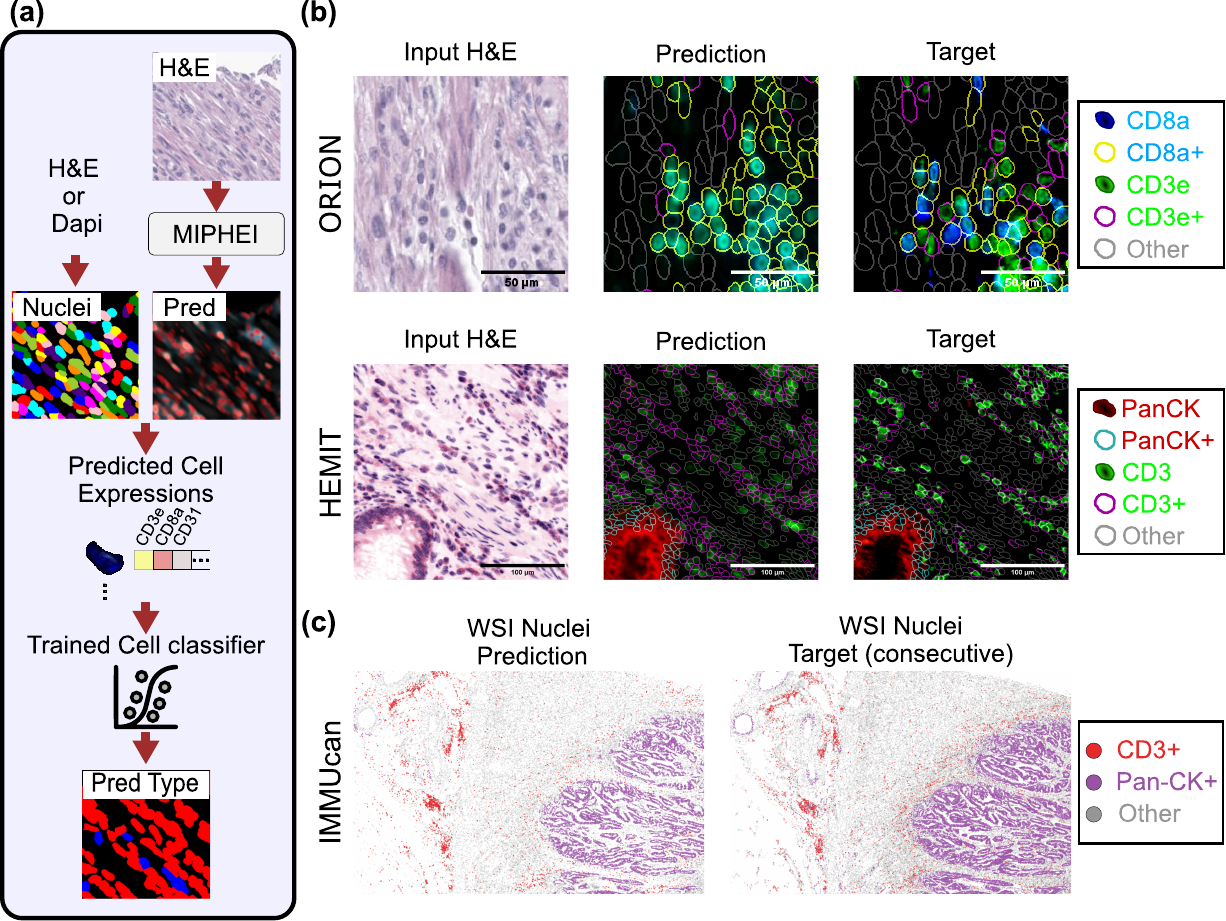}}
\caption{\textbf{Prediction pipeline and prediction visualization:} \textbf{(a) Inference pipeline:} mIF images are first generated using a trained U-Net model. Predicted single-cell data are then extracted by averaging predicted mIF signals within each nucleus, using nuclei masks from an external segmentation model. Finally, a cell classifier, trained on validation set cells, predicts cell types. \textbf{(b) Prediction examples from our best model:} Predicted mIF images and cell types (shown as colored cell boundaries) are compared to target mIF images and annotated cell types from the same restained tissue section in the OrionCRC (CD3e, CD8a) and HEMIT (Pan-CK, CD3) datasets. \textbf{(c) IMMUcan large-area visualization:} Nuclei predictions on H\&E alongside clustered nuclei from the corresponding consecutive mIF sections.}
\label{figure_vis}
\end{figure}

\subsection{Training Setup}

\subsubsection{Data Configuration}

We split the OrionCRC dataset by slide into training (37), validation (2), and test (2) sets. For HEMIT, we used its original train-validation-test splits. For IMMUcan, Pathocell, Lizard and PanNuke, all tiles were treated as test set.

All models are trained on H\&E-mIF images from OrionCRC extracted as 256x256 pixel tiles at 0.5 mpp. For normalization, target data is scaled to the range \([-0.9, 0.9]\) to prevent saturation at extreme values in Tanh activation. Input H\&E RGB data is normalized using the mean and standard deviation of the foundation model employed.

To enhance generalization, augmentations include spatial transformations, such as horizontal and vertical flips and coarse dropout (zeroing out a random box in both input and target), along with color augmentations like stain augmentation, random brightness and contrast adjustments, gaussian blur, and gaussian noise. These augmentations simulate variations in stain intensity, color distributions, and common imaging artifacts. To improve robustness, we trained a CycleGAN \citep{cyclegan} following \citet{cycleganhisto} to translate OrionCRC H\&E images toward the IMMUcan style, capturing more complex domain shift, and used both original and precomputed translated tiles as augmentation during training.

\subsubsection{Training Configuration}

We adopt similar Pix2Pix hyperparameters (excluding adversarial training), given its role as a classical image translation model. We use the Adam optimizer with a learning rate of \(2 \times 10^{-4}\), kept constant in the first half of training before linearly decaying to zero. A linear warmup phase is applied to the generator’s learning rate for the first 400 iterations to improve stability and adaptation, as recommended for ViTs by \citet{b25}. On top of data augmentations, regularization includes weight decay (\(1 \times 10^{-5}\)), gradient clipping (max norm 1), and dropout (0.1). Non-pretrained weights are initialized using a normal distribution with a gain of 0.02, and biases are set to zero. All trainings were conducted with a batch size of 16. All experiments were performed on a single A100 GPU.

\subsection{Metric Overview}
\label{subsec:metric_overview}

To evaluate our models, we report both pixel-level and cell-level metrics.

\subsubsection{Pixel-level metrics}

At the pixel level, we compute per-channel PSNR, SSIM, and Pearson correlation, and report the average across channels. Pearson correlation is computed over the full evaluation set rather than averaged per tile. As shown in \citep{isbi_benchmark}, Pearson correlation is less sensitive to background proportion than PSNR and SSIM, making it more suitable for sparse mIF markers.

\subsubsection{Cell-level metrics}
\label{subsection:cell_metrics}

A key objective of mIF prediction is to preserve biologically meaningful cell-type information for downstream TME analysis. Pixel-level metrics alone do not capture this property, so we also evaluate performance at the cell level. For each nucleus, we extract predicted single-cell expression by averaging the predicted mIF intensity within each cell instance (see Figure \ref{figure_vis}.a). These predicted cell-level expression vectors are used to train a logistic regression classifier to match cell pseudo-labels (cf Section~\ref{subsubsec:single-cell-clustering}). The same training and evaluation protocol is applied to all models: for OrionCRC, the cell classifier is trained on validation cells and evaluated on the test set, while for all other external datasets it is trained on a randomly selected 20\% subset of cells and evaluated on the remaining 80\%. Analogous to linear probing in SSL, this classifier evaluates whether predicted mIF signals preserve discriminative biological structure without relying on complex downstream models. Compared to direct gating on predicted mIF channels, this approach is more robust to individual channel errors and differences in intensity calibration between predicted and real mIF.

Following \citep{isbi_benchmark}, we report Area Under the Precision-Recall Curve (AUPRC) and F1 score between predicted cell types and cell pseudo-labels on a test set. These metrics are well suited for highly imbalanced cell populations, which are common in mIF-based cell phenotyping, with many more negative than positive cells.

To assess the statistical robustness of our cell-level metrics, we conducted a bootstrap analysis. We generated 1,000 bootstrap samples, each the same size as the original test set, by resampling tiles with replacement. A fixed random seed was used to ensure reproducibility and enable consistent comparison across models. For each sample, we computed the AUPRC and F1 scores and we then reported 95\% confidence intervals using the 2.5th and 97.5th percentiles of the resulting distributions. These intervals are presented in Figure~\ref{figure_perfs}.

We further analyze our model by computing cell count correlations on IMMUcan for CD3, CD8, CD4, FOXP3, and Pan-CK. Since consecutive tissue sections in the IMMUcan dataset are biologically similar but not identical, direct pixel-level alignment is not feasible. However, the minimal spatial separation between sections preserves overall tissue architecture and cellular distribution, allowing for meaningful region-level comparisons. We compute the correlation between predicted and target positive cell counts across registered tiles, using predictions from the logistic regression-based single-cell classification. Following preprocessing in D.1, we compute Pearson correlation coefficients between tile-level cell counts (0.26 mm$^2$ area per tile) for each marker. High correlations, even if imperfect, indicate effective cell identification.

\section{Results}

This section presents an ablation study to identify critical model components, a comparison with other methods, including HEMIT, ROSIE, DiffusionFT, and Pix2Pix, and a detailed  analysis of marker-specific performance. We exclude HistoPlexer \citep{histoplexer}, as it is designed for consecutive-section supervision, which is not relevant in our same-section evaluation setting.

\subsection{Ablation studies}

To identify critical components and configurations significantly impacting performance, we trained all models for 15 epochs ($\sim$276k iterations) with batch size 16 by default.

\begin{table}[t]
  \centering
  \footnotesize
  \begin{tabular}{@{}lccccc@{}}
    \toprule
    Configuration & \multicolumn{3}{c}{Reconstruction} & \multicolumn{2}{c}{Cell Classification} \\
    \cmidrule(lr){2-4} \cmidrule(lr){5-6}
    & PSNR & SSIM & \textbf{Pearson} & \textbf{AUPRC} & \textbf{F1} \\
    \midrule
    
    \multicolumn{6}{@{}l}{\textbf{GAN Discriminator} (MIPHEI-UNETR H-optimus-0 LoRA)} \\
    Pix2Pix (GAN) & 30.88 & 0.942 & 0.395 & 0.531 & 0.499 \\
    \textbf{Generator only} & \textbf{31.89} & \textbf{0.951} & \textbf{0.466} & \textbf{0.564} & \textbf{0.515} \\
    \addlinespace
    
    \multicolumn{6}{@{}l}{\textbf{Foundation Model Encoder} (MIPHEI-UNETR with LoRA)} \\
    CTransPath & 30.77 & 0.947 & 0.357 & 0.472 & 0.449 \\
    UNIv2 & 31.88 & 0.950 & 0.454 & 0.554 & 0.508 \\
    \textbf{H-optimus-0 (selected)} & \textbf{31.89} & \textbf{0.951} & \textbf{0.466} & \textbf{0.564} & \textbf{0.515} \\
    \addlinespace
    
    \multicolumn{6}{@{}l}{\textbf{Encoder Finetuning Method} (MIPHEI-UNETR H-optimus-0)} \\
    Frozen encoder & 31.63 & 0.949 & 0.434 & 0.541 & 0.497 \\
    \textbf{LoRA (selected)} & \textbf{31.89} & \textbf{0.951} & \textbf{0.466} & \textbf{0.564} & \textbf{0.515} \\
    \addlinespace
    
    \multicolumn{6}{@{}l}{\textbf{Impact of Architecture}} \\
    MIPHEI-ConvNeXtv2 L & 31.51 & 0.949 & 0.404 & 0.508 & 0.473 \\
    MIPHEI-HEMIT & 31.38 & 0.949 & 0.388 & 0.492 & 0.457 \\
    MIPHEI-UNETR & 31.89 & 0.951 & 0.466 & 0.564 & 0.515 \\
    \textbf{MIPHEI (Best)} & \textbf{31.93} & \textbf{0.951} & \textbf{0.470} & \textbf{0.574} & \textbf{0.523} \\
    \bottomrule
  \end{tabular}
  \caption{Ablation study on the OrionCRC test set with the optimal MIPHEI configuration shown in bold. Reconstruction performance is evaluated at the pixel level using PSNR, SSIM, and Pearson correlation. Cell-level performance is also reported using AUPRC and F1 score, as described in Section~\ref{subsection:cell_metrics}.}
  \label{tab:ablation_study}
\end{table}

\subsubsection{Impact of Discriminator}

We evaluated the impact of adding a discriminator in a Pix2Pix-like setup, using UNETR as the generator and the standard patch-based discriminator from Pix2Pix. We found that the discriminator slightly reduces pixel-level metrics, and consistently cell-type classification accuracy. We hypothesize the model favors realism over fidelity (Table \ref{tab:ablation_study}), thus leading to more realistic looking images, but not providing more accurate predictions. For this reason, we excluded the discriminator from further experiments.

\subsubsection{Architecture Comparison}
\label{subsubsec:architecturecomparison}

We compared three MIPHEI variants for mIF prediction. One uses convolutional encoder pretrained on ImageNet: ConvNeXt v2 Large (203M parameters). The other two variants use the H-optimus-0 foundation model as encoder but differ in their ViT integration strategies: UNETR (1.17B parameters, 36M trainable via LoRA) and ViTMatte (1.14B parameters, 6.6M trainable via LoRA). MIPHEI-ConvNeXt achieved strong results, benefiting from high model capacity and attention-like mechanisms. Transformer-based variants using H-optimus-0 encoder significantly outperformed the convolutional one, emphasizing the advantage of employing pretrained foundation models for the encoder component in our task. UNETR and ViTMatte achieved similar performance, but ViTMatte converged much faster, reaching UNETR's 10-epoch performance after the first one only. This is likely due to its lightweight convolutional Detail Capture module, which more directly complements ViT representations and aid spatial reconstruction compared to UNETR’s token-based upsampling. Overall, ViTMatte offers the best trade-off between performance and training efficiency and was therefore selected as the default MIPHEI architecture.

\subsubsection{Foundation Model Benchmark}
\label{subsubsec:fmbenchmark}

We evaluated three foundation-model encoders available for our image translation task (UNETR with LoRA): CTransPath, Univ2, and H-optimus-0. CTransPath was significantly outperformed by Univ2 and H-optimus-0, which provided similar results with a slight advantage for H-optimus-0, which we therefore selected.

\subsubsection{Finetuning Strategies}

Finally, we evaluated the finetuning strategies described in subsection \ref{subsec:finetuning-strats} using UNETR with the H-optimus-0 encoder. Our results show that LoRA improves performance over a frozen encoder, achieving a higher test cell AUC (0.431 vs. 0.413). Based on this, we adopt LoRA as our fine-tuning strategy. 

Based on our ablation study, the best-performing configuration for MIPHEI is ViTMatte with the H-optimus-0 encoder, adapted using LoRA, and trained without a GAN strategy, providing an optimal balance between performance, efficiency, and convergence speed.

\subsection{Benchmark with Existing Methods}

\begin{figure}[!t]
\centering
\includegraphics[width=\textwidth]{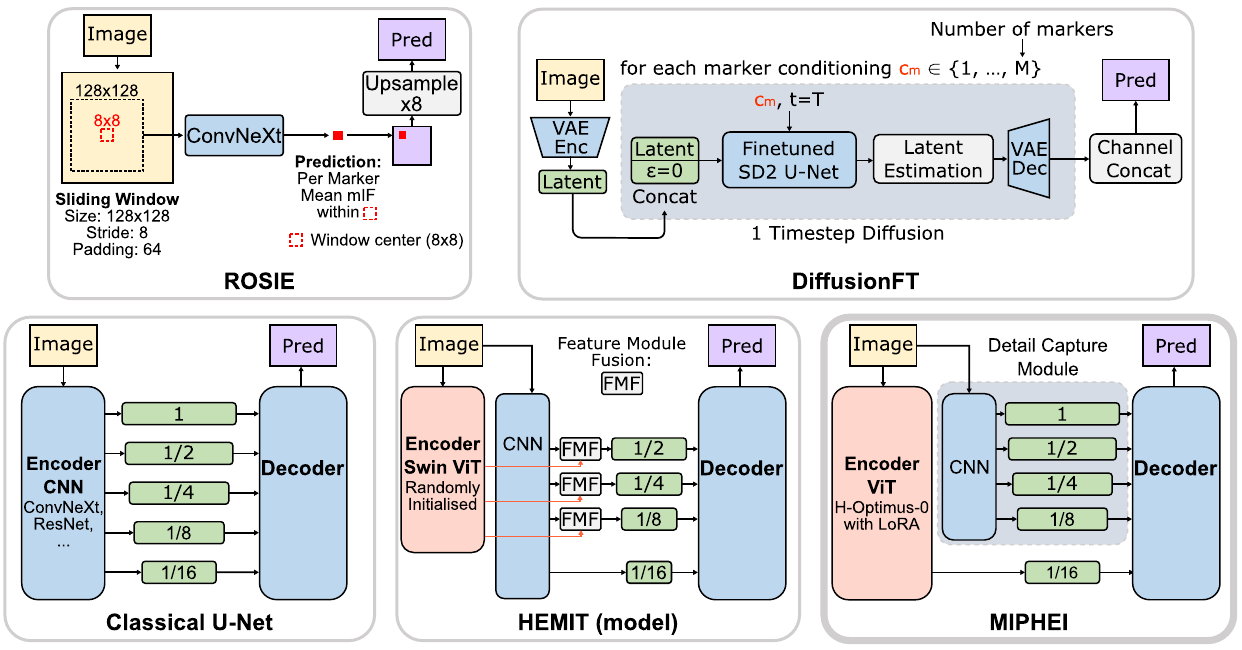}
\caption{\textbf{Architectures used in our benchmark.} Inference pipelines of the compared models, including CNN-based (top left), diffusion-based (top right), and U-Net-inspired architectures (bottom). Blue denotes convolutional blocks, yellow input images, purple predictions, and green intermediate tensors.}
\label{figure_architectures_all}
\end{figure}

To benchmark MIPHEI, against existing approaches, we evaluate in-domain and cross-dataset generalization using the following models: (1) \textbf{MIPHEI}; (2) \textbf{MIPHEI-ConvNeXt}, a variant using a ConvNeXt encoder; (3) \textbf{HEMIT}; (4) \textbf{ROSIE}; (5) \textbf{DiffusionFT}; (6) a \textbf{Nuclear Morphometry} simple baseline relying on nucleus shape and H\&E-derived statistics (e.g., area, orientation, hematoxylin intensity); (7) an \textbf{Upper Bound} model using real mIF-derived single-cell expressions, restricted to the intersection of markers available in each dataset and the predicted marker panel;(8) and a \textbf{prevalence-base random model}, which assigns cell-type probabilities according to their prevalence in the test sets, with metrics averaged over 100 runs to account for sampling variability. Since most cell subtypes are not identifiable from H\&E alone by a pathologist, the task is challenging, and the random baseline provides a reference to show that our model outperforms chance-level expectations. In constrast, the upper bound baseline estimates the performance achievable with perfectly matched predictions at the cell level.

All models are trained on the OrionCRC training and validation splits to ensure a fair comparison. An overview of the compared models is shown in Figure~\ref{figure_architectures_all}. Pix2Pix, HEMIT, MIPHEI, and ROSIE are trained using their public implementations. As the DiffusionFT model of Oh et al. is not publicly available, we adapt the open Diffusion-E2E-FT implementation and use the same hyperparameters.

\begin{figure}[!t]
\centerline{\includegraphics[width=\textwidth]{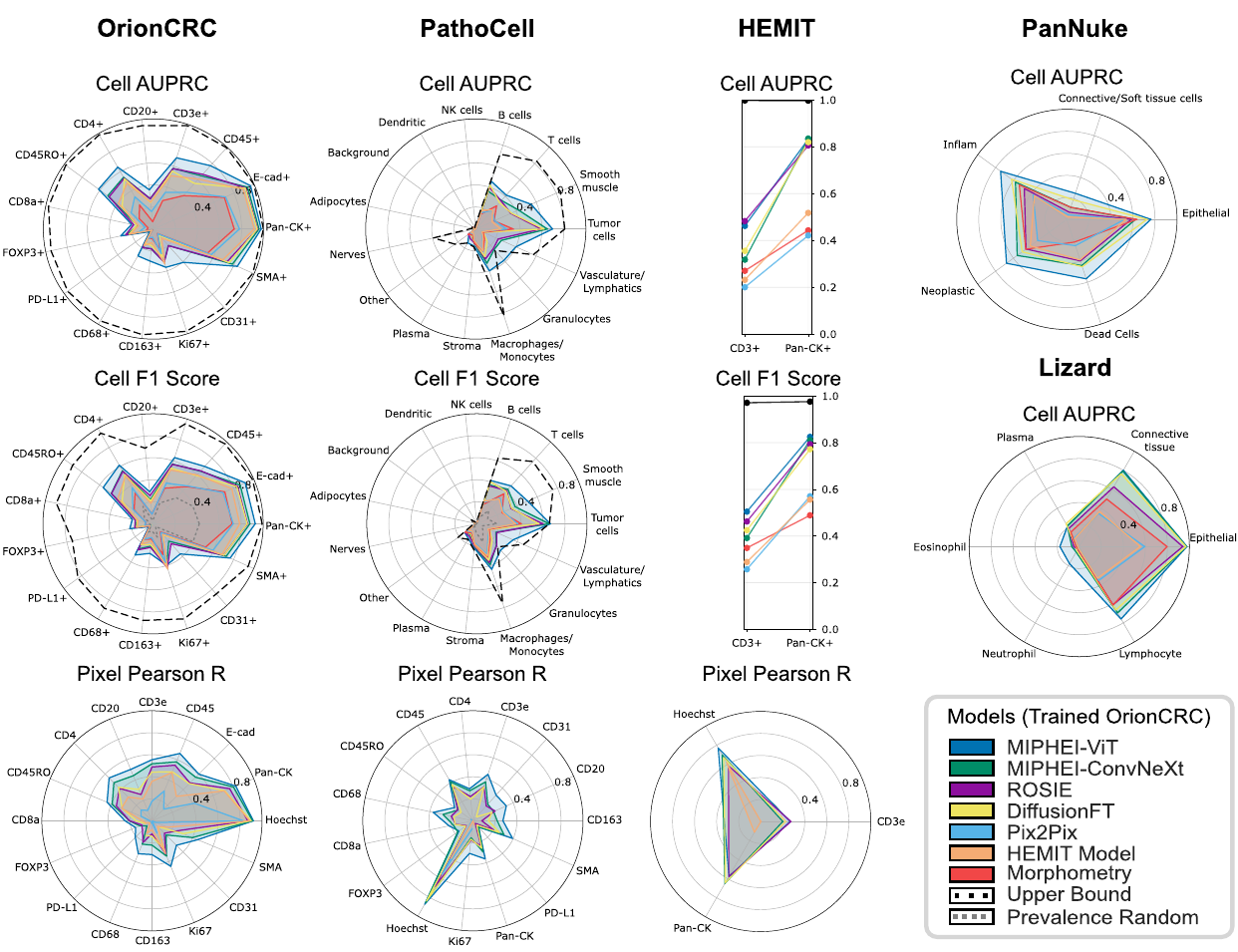}}
\caption{\textbf{H\&E-to-mIF Model Benchmark.} Comparison of H\&E-to-mIF models on OrionCRC (in-domain) and PathoCell, HEMIT, PanNuke, and Lizard (out-of-domain). We report cell-level AUPRC and F1 scores (cf Section~\ref{subsec:metric_overview}), and pixel-level Pearson correlation. All models are trained on OrionCRC. As PanNuke and Lizard do not provide mIF data, only cell-level F1 scores are reported for these datasets}
\label{figure_benchmark}
\end{figure}

Evaluation is conducted on the OrionCRC test set (in-domain), and on HEMIT, PathoCell, Lizard, and PanNuke as out-of-domain datasets. Results across models and datasets are summarized in Tables~\ref{tab:cell_auprc} and \ref{tab:pixel_pearson}, while Figure~\ref{figure_benchmark} presents the complete radar-plot comparison across datasets.

\subsubsection{Benchmark Results}

All models outperform prevalence-based random baselines, confirming that molecular information is to some degree predictable from morphology, as captured by H\&E Images. Overall, MIPHEI-ViT consistently achieves the best performance across datasets and for nearly all markers. MIPHEI-ConvNeXt, ROSIE, and DiffusionFT follow with comparable results, with their relative ranking varying across datasets. DiffusionFT shows some performance drops on immune markers such as CD45 and CD45RO but performs well on expert-annotated datasets. HEMIT achieves strong in-domain performance but a substantial performance drop out of domain. Finally, Pix2Pix and nuclear morphometry baseline consistently perform worst.

Performance generally decreases when moving from in-domain to out-of-domain datasets. MIPHEI preserves a consistent best performane across datasets, highlighting its robustness. It is likely supported by the integration of the ViT foundation encoder. MIPHEI-ConvNeXt, HEMIT, and DiffusionFT have broadly similar behavior under domain shift. GAN-based approaches, including Pix2Pix and HEMIT, are strongly affected by domain shift, often producing near-zero predictions for non-nuclear markers. Overall, a large performance drop is observed on PathoCell for all models, including the upper bound. This is expected, as cell-type labels are derived from the full 58-marker panel while evaluation is restricted to the OrionCRC marker subset, which is not sufficient to recover all cell types.

In our ablation studies, we showed that MIPHEI-HEMIT, a variant using the HEMIT generator trained within the MIPHEI pipeline (no GAN, weighted MSE), achieves an overall cell-level AUPRC of 0.49 on the OrionCRC test set. Compared to the original HEMIT trained on the same data, this represents a substantial improvement of +0.10 AUPRC on the same split. This result demonstrates the effectiveness of our training pipeline, even when applied to an alternative architecture.

\begin{table}[t]
  \centering
  \footnotesize
  \setlength{\tabcolsep}{4pt}
  \begin{tabular}{l c c c c c}
    \toprule
    \thead{Model}
      & \thead{OrionCRC}
      & \thead{PathoCell}
      & \thead{HEMIT}
      & \thead{Lizard}
      & \thead{PanNuke} \\
    \midrule

    Nuclear Morphometry
      & 0.247 & 0.111 & 0.357 & 0.352 & 0.393 \\

    Pix2Pix
      & 0.287 & 0.097 & 0.311 & 0.235 & 0.305 \\

    HEMIT
      & 0.390 & 0.094 & 0.375 & 0.227 & 0.345 \\

    ROSIE
      & 0.433 & 0.187 & 0.644 & 0.408 & 0.397 \\

    DiffusionFT
      & 0.392 & 0.169 & 0.588 & 0.451 & 0.498 \\

    MIPHEI-ConvNeXt (Ours)
      & 0.437 & 0.197 & 0.577 & 0.467 & 0.462 \\

    MIPHEI-ViT (Ours)
      & \textbf{0.517} & \textbf{0.250} & \textbf{0.647} & \textbf{0.517} & \textbf{0.60} \\
    \bottomrule
    Upper Bound
      & 0.970 & 0.396 & 0.998 & -- & -- \\
    \bottomrule
  \end{tabular}
  \caption{Cell-level performance (AUPRC) across datasets and methods. Reported values correspond to AUPRC averaged across all markers and cell-level evaluation follows the protocol described in Section~\ref{subsection:cell_metrics}. MIPHEI achieves the highest AUPRC across all datasets. Dashes indicate metrics that are not applicable.}
  \label{tab:cell_auprc}
\end{table}

\begin{table}[t]
  \centering
  \footnotesize
  \setlength{\tabcolsep}{4pt}
  \begin{tabular}{l c c c}
    \toprule
    \thead{Model}
      & \thead{OrionCRC}
      & \thead{PathoCell}
      & \thead{HEMIT} \\
    \midrule

    Pix2Pix
      & 0.153 & 0.038 & 0.141 \\

    HEMIT
      & 0.280 & 0.028 & 0.245 \\

    ROSIE
      & 0.354 & 0.222 & 0.479 \\

    DiffusionFT
      & 0.324 & 0.228 & 0.513 \\

    MIPHEI-ConvNeXt (Ours)
      & 0.404 & 0.278 & 0.517 \\

    \textbf{MIPHEI-ViT} (Ours)
      & \textbf{0.470} & \textbf{0.336} & \textbf{0.538} \\
    \bottomrule
  \end{tabular}
  \caption{Pixel-level performance (Pearson correlation) across datasets and methods. Pearson correlation is computed over the full evaluation set (not averaged per tile) and averaged across all markers. MIPHEI achieves the highest Pearson correlation across datasets. Dashes indicate metrics that are not applicable.}
  \label{tab:pixel_pearson}
\end{table}

\subsubsection{Efficiency analysis.}

Virtual staining typically requires running models over entire WSIs, making inference efficiency a critical factor. To inference compare computational cost, we report parameter counts, FLOPs, and peak GPU memory usage for a single 256×256 tile (Table~\ref{tab:efficiency}). ROSIE is the most computationally demanding due to its sliding-window inference strategy. DiffusionFT is also costly, as the diffusion model is applied once per marker, even if a single inference timestep is used. MIPHEI-ViT is heavier than CNN-based models due to its ViT-H backbone, 
but thanks to LoRA, training involves only a small number of parameters.
Pix2Pix and HEMIT are the most lightweight models but achieve lower accuracy. Overall, MIPHEI-ConvNeXt offers the best trade-off between efficiency and performance.

\begin{table}[h]
\label{tab:efficiency}
\centering
\small
\begin{tabular}{lccc}
\hline
\textbf{Model} & \textbf{Params (M)} & \textbf{FLOPs (G)} & \textbf{VRAM (GiB)} \\
\hline
Pix2Pix            & 54.4   & 6.49    & 0.23 \\
HEMIT              & 79.7   & 65.17   & 0.40 \\
ROSIE              & 49.5   & 8910.23    & 0.21 \\
MIPHEI-ViT         & 1141.6$^{*}$ & 383.68 & 4.47 \\
MIPHEI-ConvNeXt    & 203.3  & 49.49   & 0.82 \\
DiffusionFT        & 951.3  & 6490.07    & 4.19 \\
\hline
\end{tabular}
\caption{
Model inference computational summary.
All results are reported for a $256 \times 256$ input image using float32.
Parameters are shown in millions (M), FLOPs in gigaFLOPs (G), and VRAM in GiB.
$^{*}$Only 6.7M parameters are trainable.
}
\end{table}

\subsection{Marker-Level Analysis} 

\begin{figure}[!t]
\centerline{\includegraphics[width=\textwidth]{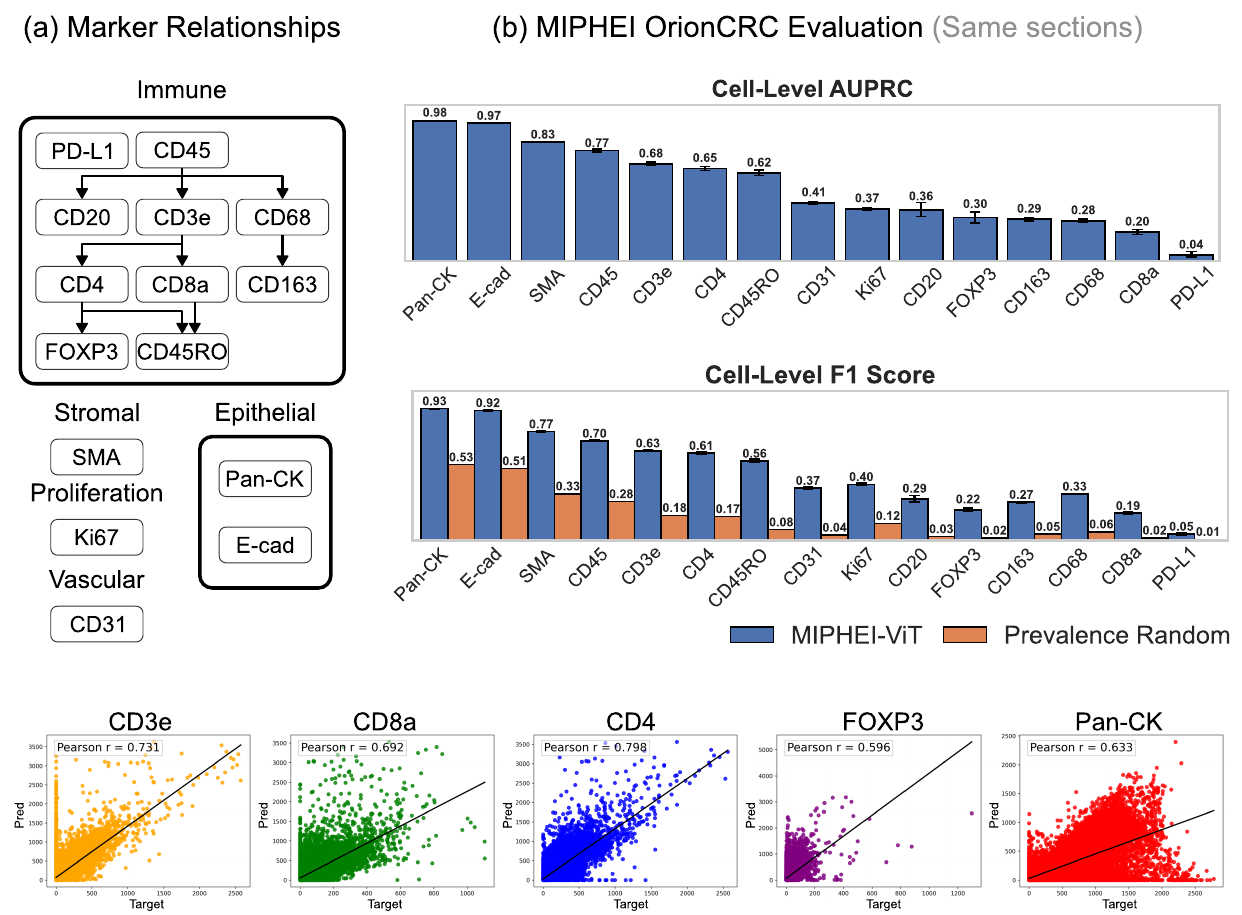}}
\caption{\textbf{MIPHEI Evaluation analysis:} 
\textbf{(a) Marker associations} and phenotypic groupings used to interpret differences in marker predictability. Arrows denote dominant subset relationships between phenotypes (e.g., CD3e+ cells largely co-express CD45). \textbf{(b) Cell-level performance of MIPHEI on OrionCRC.} Mean predicted marker expression is extracted per cell and used to train a logistic regression classifier to match pseudo cell types on validation cells, then evaluated on test cells. We report cell-level AUPRC and F1 scores, with a prevalence-based random baseline for F1 to assess whether meaningful biological patterns are learned. Error bars indicate 95\% confidence intervals obtained by bootstrapping (Section~\ref{subsec:metric_overview}). \textbf{(c) IMMUcan evaluation of MIPHEI on consecutive sections.} Pearson correlation and linear regression plots between predicted and pseudo-labeled cell-type counts across 17k tiles.}
\label{figure_perfs}
\end{figure}

In this section, we focus on the analysis of MIPHEI. Results are illustrated in Figure~\ref{figure_perfs}.

\subsubsection{In-domain performance analysis}

We have demonstrated that MIPHEI consistently outperforms random predictions for all markers, confirming its ability to capture meaningful morphological cues from H\&E to estimate protein expression. 
However, the predictability varies across markers and cell types (Figure \ref{figure_perfs}.a).

Markers with the highest performance include epithelial markers E-cadherin (F1 0.92) and Pan-CK (F1 0.93), which label epithelial cells forming well-defined clusters and glandular structures in H\&E. The immune marker CD45 (F1 0.70) also performs well. While it mainly identifies lymphocytes, which are easily recognizable in H\&E, it also includes cells like monocytes, which are harder to spot, making prediction a bit more challenging. $\alpha$-SMA also achieves strong performance (F1 0.77), as it labels stromal and vascular structures that can be identifiable on H\&E.

Markers with moderate performance included CD31 (F1 0.37), which mark endothelial cells forming vascular structures visible on H\&E. However, these cells represent a relatively small population at the cell level, which makes accurate cell-level prediction more challenging. Immune subtype markers CD3e, CD45RO, CD4 and CD20 (F1 0.29–0.63) also show moderate performance, with broader T-cell markers like CD3e (F1 0.63) achieving better performance than more specific ones like CD45RO (F1 0.56). CD20, a B-cell marker, shows lower performance within this group, likely due to the more limited representation of B cells in OrionCRC. While lymphocytes are visible in H\&E, their subtypes remain indistinguishable to pathologists, highlighting our model’s value on this difficult task. Macrophage markers CD68 (F1 0.362) and CD163 (F1 0.206) faced similar challenges, as their heterogeneity complicates identification, with CD163 being a macrophage subtype of CD68, further complicating prediction. Ki-67 also exhibits moderate performance, as it labels proliferating cells across the G1, S, and G2 phases that lack distinctive morphological features on H\&E, with only a small fraction corresponding to visible mitotic figures. Finally, CD68 shows moderate performance, likely due to the heterogeneous morphology of macrophages and the diffuse nature of its expression.

Markers with the lowest performance are CD163 (F1 0.27), FOXP3 (F1 0.22), CD8a (F1 0.19), and PD-L1 (F1 0.05). CD163 shows limited performance because it marks M2 macrophages, a functional subset of CD68+ cells, and H\&E probably does not allow to distinguish macrophage activation states. FOXP3 labels rare regulatory T-cells, a highly specific CD4+ subset, making it one of the most difficult immune markers to predict from morphology alone. Similarly, CD8a marks a less abundant T-cell subtype that is typically more dispersed than CD4+ cells. Finally, PD-L1 is a functional marker that is rarely expressed and not associated with a specific cell type, and shows low-intensity, noisy staining in OrionCRC, making prediction from H\&E particularly challenging.

In summary, epithelial  markers, broad immune markers and $\alpha$-SMA seem predictable with high accuracy. Markers defining specific immune subtypes are harder to predict, and some functional markers are not predictable with an accuracy high enough to be applicable in practice. Overall, prediction performance decreases as markers correspond to increasingly rare or specialized cell populations.

\subsubsection{Out-of-domain performance analysis}

Our external validation on HEMIT, PathoCell and IMMUcan confirms that MIPHEI, trained on OrionCRC, generalizes well to datasets with domain shifts in mIF technology and H\&E appearance, requiring minimal adaptation.

On the HEMIT dataset, analysis is limited by the small number of available markers (CD3e and Pan-CK), yet MIPHEI shows good performance with minimal adaptation. These results should be interpreted with caution, as tiles were selected from regions containing many CD3+ lymphocytes and Pan-CK+ cells, which simplifies the task and may inflate performance. In this setting, a model predicting all lymphocytes as T cells would already perform well. At the same time, stain normalization applied to all H\&E images introduces a strong, artificial domain shift.

On PathoCell, overall performance is moderate, including for the upper bound, due to the mismatch between the full 58-marker PathoCell panel used for labeling and the restricted set of predicted markers, as discussed earlier. Despite this limitation, we observe trends consistent with the in-domain analysis, with the best performance for T cells, B cells, smooth muscle, and tumor cells, which are directly linked to the predicted protein markers. Although absolute scores decrease, the gap relative to the upper bound remains similar. MIPHEI even outperforms the upper bound for granulocytes, likely because the predicted panel lacks a granulocyte-specific marker and the models leverage morphological proxies. At the pixel level, Hoechst reconstruction remains strong but other marker degrades compared to OrionCRC. The absence of autofluorescence subtraction, which slightly shifts the target signal, may explain this behavior.

\begin{figure}[!t]
\centerline{\includegraphics[width=\textwidth]{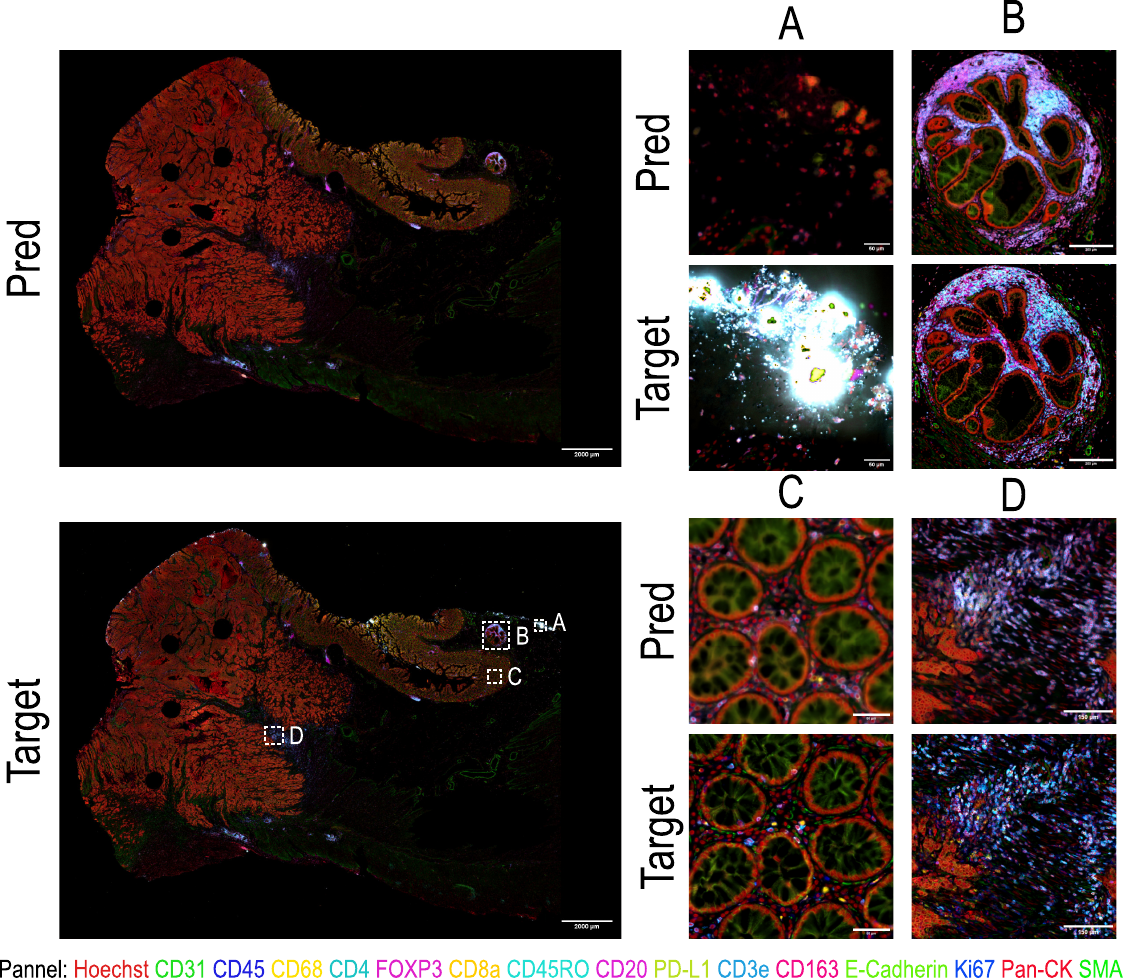}}
\caption{\textbf{WSI prediction of mIF from H\&E using MIPHEI-ViT.} 
Top left: MIPHEI-ViT prediction overlay (RGB composite of 16 mIF markers) for a full CRC WSI. 
Bottom left: corresponding real mIF target. 
Right: zoom-in on four selected regions \textbf{(A–D)} with model predictions (top) and ground truth (bottom). 
The regions cover diverse tissue contexts, including an artifact in \textbf{(A)} that is correctly ignored by the model. (see \textit{Supplementary Video 2} for an interactive side-by-side demonstration in QuPath).}
\label{figure_wsi_pred}
\end{figure}

We further evaluate model generalization on the IMMUcan dataset by correlating cell-type counts predicted by MIPHEI from H\&E with pseudo-label counts obtained from consecutive mIF sections. Pearson correlation is computed over 0.26 mm$^2$ regions using a logistic regression cell classifier trained on OrionCRC validation cells with MIPHEI single-cell predictions. We observe strong correlations for CD4 (0.80), CD3e (0.73), and CD8 (0.69), and moderate correlations for FOXP3 (0.60) and Pan-CK (0.63). We also observe overdetection for CD3e (i.e. cells with CD3e prediction and measured 0 expression) and underdetection for Pan-CK  (cells with predicted 0-expression and positive according to the measurement). Manual inspection showed that this was mainly due to low-quality tiles with high auto-fluorescence. In Figure~\ref{figure_wsi_pred}.A, we observe a similar case on another dataset, with an artifact present in the target but correctly not predicted by the model.

On Lizard, performance is good for epithelial cells, and remains acceptable for connective tissue ($\alpha$-SMA) and lymphocytes. Plasma cells and neutrophils are weakly supported by the predicted marker panel and therefore show low performance. Eosinophils, which are not specifically covered by the panel, exhibit very low performance. On PanNuke, the cross-organ setting and higher magnification introduce a stronger domain shift, leading to overall lower performance. Neoplastic and dead cell classes show more limited performance, as their identification relies on indirect cues from the predicted panel. Neoplastic cells can be inferred from Pan-CK, E-cadherin, and Ki67, while dead cells are characterized by low or absent signal across all predicted markers. Unexpectedly, connective tissue performance is also limited, which may be partly explained by the very small number of positive cells.

\section{Discussion}

In this study, we present MIPHEI, a method trained to predict 16 mIF channels from standard H\&E slides. For this, we proposed a U-Net with a ViT-based foundation model as encoder and demonstrated that this architecture outperforms previously proposed methods. Moreover, we showed that MIPHEI generalizes well across datasets. We attribute this to the robust and transferable encodings learned by the foundation model, which was exposed to millions of tiles during pretraining. This contrasts with prior reports \citep{histoplexer,rosie}, where foundation models underperformed, likely due to limited fine-tuning strategies or insufficient integration within the architecture. Our findings highlight the value of leveraging foundation models for mIF prediction, and potentially for other image translation tasks in histology.

We also introduced a validation strategy focused on single-cell metrics, which we believe are the most relevant for this task. Pixel-wise accuracy may be unreliable, as H\&E images are usually not informative about cytoplasmic boundaries, and mIF targets are often sparse. Instead, the biologically meaningful information lies in protein expression levels within individual cells, protein positivity, or the resulting cell type. We reflect this in our evaluation framework, which we provide as part of this study.

Accurate, domain-robust mIF prediction opens the door to a range of applications. While direct clinical deployment for diagnostics may remain challenging, MIPHEI proves to be a powerful tool for mining large retrospective cohorts, enabling the identification of cell types without relying on manual annotation. Ultimately, this can allow to identify associations between specific cell populations, their spatial arrangements, and clinically relevant outcomes such as survival or treatment response. MIPHEI thus holds promise for hypothesis generation and exploratory analysis. Furthermore, extending this approach to predict clinically relevant scores, such as the Immunoscore, represents a compelling direction for future work.

Our study is not free of limitations. Although the number of training tiles was substantial, they were derived from only 41 slides. Access to larger and more diverse datasets would likely improve the model's robustness to domain shifts and biological variability. Additionally, we observed that the cell classification model still requires fine-tuning on a small set of labeled cells when applied to out-of-domain data. As such, if a domain shift is anticipated, some mIF data will still be needed for calibration.

\section{Conclusion}
In this paper, we present MIPHEI, a deep learning framework that predicts mIF images from H\&E using ViT foundation models as encoders within a U-Net architecture. MIPHEI outperforms state-of-the-art models on both internal and external datasets, demonstrating strong generalization across staining protocols and imaging conditions. We evaluated MIPHEI across 15 protein markers and associated cell types. On our in-domain test set, it achieves high accuracy for epithelial (Pan-CK, E-cadherin), broad immune markers (CD45, CD3e) and stromal markers ($\alpha$-SMA), performs moderately on more specific immune subtypes (CD4, CD45RO, CD20, CD68) and vascular markers (CD31), and struggles with CD8a, FOXP3 and PD-L1 due to their complexity and small number of positive cells.

\section*{CRediT authorship contribution statement}

\textbf{Guillaume Balezo:} Conceptualization, Data curation, Formal analysis, Investigation, Methodology, Software, Validation, Visualization, Writing – original draft, Writing – review \& editing. \textbf{Roger Trullo:} Conceptualization, Funding acquisition, Methodology, Project administration, Resources, Supervision, Writing – review \& editing. \textbf{Albert Pla Planas:} Conceptualization, Methodology, Project administration, Resources, Supervision, Visualization, Writing – review \& editing. \textbf{Etienne Decenciere:} Conceptualization, Methodology, Project administration, Supervision, Visualization, Writing – review \& editing. \textbf{Thomas Walter:} Conceptualization, Funding acquisition, Methodology, Project administration, Supervision, Visualization, Writing – original draft, Writing – review \& editing.

\section*{Declaration of competing interest}
G. Balezo, R. Trullo, A. Pla Planas are/or were Sanofi employees and may hold shares and/or stock options in the company. E. Decencière and T. Walter have nothing to disclose.

\section*{Acknowledgment}

This work uses IMMUcan data funded by IMI2 JU (grant 821558) with support from Horizon 2020 and EFPIA.

This work was sponsored by ANRT and Sanofi. Furthermore, T. Walter acknowledges funding from Agence Nationale de la Recherche under the France 2030 program, with the reference numbers ANR-24-EXCI-0001, ANR-24-EXCI-0002, ANR-24-EXCI-0003, ANR-24-EXCI-0004, ANR-24-EXCI-0005, as well as funding by the French government under the management of Agence Nationale de la Recherche as part of the “Investissements d’avenir” program, reference ANR-19-P3IA-0001 (PRAIRIE 3IA Institute) and ANR-23-IACL-0008 (PRAIRIE-PSAI).

\section*{Data availability}

The dataset used in this study, derived from the OrionCRC and HEMIT datasets, has been made publicly available on Zenodo at \url{https://doi.org/10.5281/zenodo.15340874}~\citep{miphei_dataset}. Only the IMMUcan data used for validation is not redistributed, as it remains private and subject to usage restrictions.

\section*{Declaration of generative AI and AI-assisted technologies in the writing process}
We used ChatGPT (OpenAI) exclusively to improve the English language quality of the manuscript—specifically for grammar, spelling, and phrasing. It was also used to assist with formatting and documentation of the experimental code. All content was written by the authors, and the final text was carefully reviewed to ensure accuracy and clarity.

\newpage
\bibliographystyle{unsrtnat}
\bibliography{references}

\end{document}